# Ransomware detection using stacked autoencoder for feature selection

Mike Nkongolo and Mahmut Tokmak

University of Pretoria; Mehmet Akif Ersoy University, Turkey
mike.wankongolo@up.ac.za
mahmuttokmak@mehmetakif.edu.tr

**Abstract**. In response to the escalating malware threats, we propose an advanced ransomware detection and classification method. Our approach combines a Stacked Autoencoder (SAE) for precise feature selection with a Long Short-Term Memory (LSTM) classifier to significantly enhance ransomware stratification accuracy. The process involves thorough pre-processing of the UGRansome dataset and training an unsupervised SAE for optimal feature selection, or fine-tuning via supervised learning to elevate the LSTM model's classification capabilities. We meticulously analyzed the autoencoder's learned weights and activations to pinpoint essential features for distinguishing ransomware families from other malware and created a streamlined feature set for precise classification. Our results demonstrate the SAE-LSTM model's outstanding performance across all ransomware families and boasted high precision, recall, and F1 score values which underscore its robust classification capabilities. Furthermore, balanced average scores affirm the proposed model ability to generalize effectively across various malware types. To optimize the model, we conducted extensive experiments, including up to 400 epochs and varying learning rates and achieved an exceptional 99% accuracy in ransomware classification. Our proposed SAE-LSTM model surpasses the Extreme Gradient Boosting (XGBoost) algorithm, primarily owing to its effective SAE feature selection mechanism. The proposed model demonstrates outstanding performance in identifying signature attacks, and achieved a 98% accuracy rate. This result showcased its proficiency in recognizing well established ransomware patterns. In addition, a prediction of the ransomware financial impact reveals that while Locky, SamSam, and WannaCry still incur substantial cumulative costs, their individual attacks may not be as financially damaging as those of NoobCrypt, DMALocker, and EDA2.

## 1 Introduction

In today's digital age, ransomware has emerged as a significant threat to individuals and businesses alike [1]. Defined as a type of malicious software that encrypts valuable data and demands a ransom in exchange for its release, ransomware attacks have become increasingly prevalent and financially damaging [1,2]. Recent incidents have resulted in staggering losses, reaching tens of millions of dollars for consumers [3]. In June 2022, the Serbian Republic Geodetic Authority, responsible for registering property rights, experienced a ransomware attack. This attack disrupted regular services, making it difficult for citizens to make changes to real estate ownership in the registry [3].

Similar attacks have also been reported in neighboring countries. These include the Ministry of Agriculture and the Ministry of Science and Education of the Republic of North Macedonia, the Parliament, and the Council of Ministers of Bosnia and Herzegovina, various public institutions in Albania, and the majority of the governmental IT infrastructure in Montenegro [3].

South Africa, on the African continent, stands out as the country most impacted by ransomware and phishing emails [4]. The cybersecurity landscape in South Africa has exposed vulnerabilities in multiple sectors, resulting in a significant number of cyberattacks. Pieterse [5] highlights that public and private enterprises, as well as municipalities, are commonly targeted by ransomware attacks in South Africa. An example of this is the Department of Justice, which experienced its third ransomware attack in 2023, following a previous incident in 2020 [6].

These attacks have resulted in significant financial losses for various South African companies. The urgency of tackling the global problem of classifying and detecting ransomware is evident, especially when considering the security of critical infrastructure [7]. There are several different types and variants of ransomware, each with its own characteristics and behaviors (see Table 1). Nonetheless, the absence of readily accessible ransomware datasets within the current realm of intrusion detection poses a significant challenge to their accurate categorization and detection [8].

To address this limitation, we made use of the UGRansome dataset, a publicly accessible dataset created in [9], specifically designed to classify and understand ransomware [10–13]. In the age of big data, one crucial aspect of modern data analysis and machine learning is the extraction of meaningful and representative features from complex, high dimensional datasets [14]. Among the various techniques available, stacked autoencoders (SAEs) have emerged as a potent tool for automating feature discovery [14,15]. They enable the uncovering of intricate data structures and patterns. Grounded in the field of deep learning (DL), SAEs provide an effective solution for addressing the challenge of representing high dimensional data. They pave the way for improved predictive modeling, efficient dimensionality reduction, and insightful data interpretation [15].

## 2   Stacked AutoEncoder Background

Feature selection and extraction using SAEs have been extensively studied in various domains. Wang et al. [18] proposed the use of Broad Autoencoder Features (BAF), which involves four interconnected SAEs with different activation functions. The study proposes the BAF with four parallel connected SAEs using different activation functions and evaluates the performance of the BAF in terms of learned features using the Deep Neural Network (DNN). Another study by Kong et al. [19] explored the topic of feature extraction of load curves using an autoencoder network. Wang et al. [20] used a Stacked Supervised Auto-Encoder (SSAE) to train the deep network to obtain fault relevant features. By stacking multiple supervised autoencoders, high level fault relevant features are learned to improve the classification accuracy. In [21] the integration of SAE characteristics with wavelet-based and morphological fractal texture attributes was proposed for the classification of skin disorders. This approach achieved high accuracy in the classification task. Kim et al. [22] focused on proposing an SAE-based Convolutional Neural Network (CNN) model using discrete wavelet transform for feature extraction. The model aims to improve the accuracy of tool condition diagnosis by incorporating features from cutting force data, current signal, and coefficients of the discrete wavelet transform. In a paper by [23] a deep learning architecture with SAEs for intelligent malware detection based on Windows Application Programming Interface (API) was proposed. Similarly, [24] analyzed the effectiveness of various deep learning and machine learning classifiers in detecting Android malware applications. The study uses different datasets and explores the use of Gabor filters and autoencoders to enhance classifier performance. In [25] a novel ensemble model, called Stacked Ensemble—Autoencoder (SEAE) for malware detection in the Internet of Things (IoT) domain was developed. The proposed model utilizes three lightweight neural network models trained on essential features extracted from the *MalIng* dataset. The model demonstrates high accuracy (99.43%) in classifying malware images, outperforming existing approaches. Overall, these studies emphasize the benefits of using SAEs for feature selection and extraction across different domains and tasks.

Table 1: Types of Ransomware

| Ransomware Type | Description |
|---|---|
| Crypto | This type of ransomware encrypts the victim's files, making them inaccessible. Victims are then presented with a ransom demand. |
| Locker | Locker ransomware does not encrypt files but locks users out of their system or device. |
| Scareware | Scareware displays fake security alerts or warnings, often claiming that the victim's computer is infected with malware or illegal content. Users are tricked into paying a fee for bogus security software or services. |
| Mobile | Targeted at mobile devices, this ransomware type can lock a smartphone or tablet, encrypt files, or display threatening messages demanding payment. |
| Ransomware-as-a-Service | RaaS allows cybercriminals to rent or purchase ransomware tools. |
| Doxware or Leakware | Doxware threatens to leak sensitive or confidential data unless the ransom is paid. This adds an extra layer of pressure on victims to comply. |
| Cerber Ransomware | Cerber is a notorious ransomware family known for its ability to evade detection and rapidly evolve. It has been responsible for a significant number of attacks. |
| WannaCry Ransomware | WannaCry gained worldwide attention in 2017 when it infected hundreds of thousands of computers [16]. It exploited a Windows vulnerability to spread. |
| Ryuk Ransomware | Ryuk is a targeted ransomware strain that primarily targets businesses and organizations. It often demands large ransoms. |
| NotPetya Ransomware | This ransomware variant, which emerged in 2017 [17], was initially disguised as a ransomware attack but was later revealed to be a destructive wiper malware. |

## 3    LSTM Background

Our research introduces a unique approach that combines feature selection using SAE and classification with Long Short-Term Memory (LSTM). This resulted in improved ransomware classification accuracy. The process includes preprocessing the UGRansome dataset, training an unsupervised SAE for feature extraction, and then fine-tuning the LSTM model with supervised learning to enhance its classification capabilities.

LSTM is a type of Recurrent Neural Network (RNN) architecture that is used for processing sequential data [26]. Unlike traditional RNNs, LSTM is designed to capture long-term dependencies effectively. It achieves this by using a memory cell, which has three components: an **input gate**, a **forget gate**, and an **output gate**. The input gate decides how much new information should be stored in the memory cell, while the forget gate determines what information should be forgotten. The output gate controls the amount of information that is outputted from the memory cell to the next step.

Using these gates, LSTM can process sequential data more accurately and capture long term dependencies [26]. LSTM networks have demonstrated considerable promise in the field of detecting malware. Researchers have invested substantial effort into optimizing LSTM hyperparameters specifically for the design of Intrusion Detection Systems (IDS) [26, 27]. These endeavors have led to the exploration of various LSTM configurations and revealed that the importance of hyperparameters for LSTM in IDS differs significantly from their roles in language models.

The intricate interplay between these hyperparameters has a pronounced impact on their relative significance. Taking this interplay into account, the hierarchy of importance for LSTMs in IDS becomes clear, with batch size emerging as the most critical factor, followed by dropout ratio and padding [26]. Additionally, innovative sensitivity-based LSTM models have been proposed for the creation of System-call Behavioral Language (SBL) models for malware detection [27]. These models have demonstrated impressive performance metrics, including high accuracy values and specificity when tested on unfamiliar IDS datasets. Another approach involves leveraging LSTM in conjunction with word embedding and attention mechanisms to effectively represent and classify malware files [26]. This strategy has yielded remarkable results, achieving high accuracy and F1 scores [27]. Fang et al. [28] conducted a study where they introduced a novel method for zero-day detection using LSTM. Their model is designed specifically for identifying malicious JavaScript code injected into web pages [29] by extracting features from the semantic level of bytecode and optimizing word vectorization techniques. The findings of their research revealed that the LSTM-based detection model outperforms existing models that rely on Tree-based algorithms. In addition, Roberts and Nair [30] propose a neural architecture that addresses the problem of anomaly detection in discrete sequence datasets. Their approach involves modifying the LSTM autoencoder and incorporating an array of one-class support vector machines (SVM) to detect anomalies within sequences. This method demonstrates improved stability and performs better compared to traditional LSTM-based and sliding window anomaly detection systems. One limitation of this approach is that it requires a labeled dataset for training the one-class SVM, which can be challenging to obtain in certain domains.

## 4    Research Contribution

Our research endeavors to harness the combined power of SAEs and LSTM networks to enhance the classification and detection of ransomware using the UGRansome dataset. Specifically, the focus is on incorporating feature selection techniques within the SAE architecture to facilitate the extraction of the most relevant and discriminative features from ransomware data. By selecting the input data, the subsequent LSTM network can efficiently capture the temporal relationships within the feature space. The ultimate goal of this study is to contribute to the advancement of proactive and robust ransomware recognition and classification strategies. The approach employed in this research holds several key advantages for enhancing cybersecurity, particularly in the realm of ransomware detection. The subsequent sections of this work delve into the methodology, experimental setup, results, and discussions, all of which culminate in a comprehensive analysis of the proposed SAE-based ransomware classification using the LSTM model.

## 5    Methodology

In 2021, Nkongolo et al. [9] introduced the UGRansome dataset as a valuable resource for detecting ransomware attacks, including zero-day threats [11, 31]. Unlike other datasets in the field of IDS, UGRansome includes previously unexplored types of ransomware attacks [11,32]. It covers a range of malware categories, such as **Signature (S)**, **Anomaly (A)**, and **Synthetic Signature (SS)**, with labeled instances of well-known ransomware variants like Locky, CryptoLocker, WannaCry, as well as advanced persistent threats (APT) [8,31].

To understand the dataset in more detail, we refer to Table 2, which highlights its attribute characteristics.

Table 2: Attributes of the UGRansome Dataset

| Column | Attribute | Meaning | Type | Example 11 |
|---|---|---|---|---|
|  | Time | Column with integers indicating the timestamp of network attacks. | Numeric | 50s |
| 2 | Protocol | Column representing the network protocol used. | Categorical | TCP, ICMP |
| 3 | Flag | Column indicating network connection. | Categorical | SYN, ACK |
| 4 | Family | Column describing the ransomware category. | Categorical | WannaCry |
| 5 | Clusters | Column with integers denoting malware clusters or groups. | Numeric | 1-12 |
| 6 | SeedAddress | Column representing formatted ransomware attack links. | Categorical | 18y345 |
| 7 | ExpAddress | Column indicating original ransomware attack links. | Categorical | y7635d |
| 8 | BTC | Column with values related to Bitcoin transactions in ransomware. | Numeric | 90.0 |
| 9 | USD | Column indicating financial damages in USD caused by ransomware. | Numeric | 32,465 |
| 10 | Netflow Bytes | Column with integers showing bytes. | Numeric | 6578 |
| 11 | IPaddress | Column with IP addresses associated with network events. | Categorical | Class A, B, and C |
| 12 | Threats | Column representing the nature of mal- | Categorical | Spam or Blacklist |
| 13 | Port | Column indicating network port. | Numeric | 5062 |
| 14 | Prediction | Column indicating the target variable. | Categorical | Anomaly (A) |

## 5.1  Stacked Autoencoder and Feature Selection

Stacked Autoencoders (SAEs) are a versatile type of neural network architecture utilized for feature extraction and dimensionality reduction in various domains. They have found applications in biometrics recognition, image recognition, natural language processing, and automatic speech recognition [33]. The stacked nature of SAEs arises from their composition, which includes multiple layers of autoencoders. Each layer is tasked with reconstructing the output of the preceding layer. Training SAEs involves two critical steps: **unsupervised pre-training** and **supervised fine-tuning** [33]. In the unsupervised pre-training phase, individual layers within the network are trained using autoencoders, which specialize in learning internal data representations. These representations serve to initialize the network weights and enhance its generalization capabilities. Subsequently, in the supervised fine-tuning stage, the pre-trained layers are assembled and jointly trained using labeled data. This approach consistently achieves exceptional accuracy rates [33].

5.2    Data Weighting Techniques

SAEs can be further enhanced by integrating data weighting techniques, which bolster the network's robustness and discriminative capacity [33]. Stacked sparse autoencoders have emerged as a powerful tool for dimensionality reduction and classifiers in intrusion detection systems [34].

5.3    SAE Architecture

Figure 1 illustrates a typical SAE architecture.

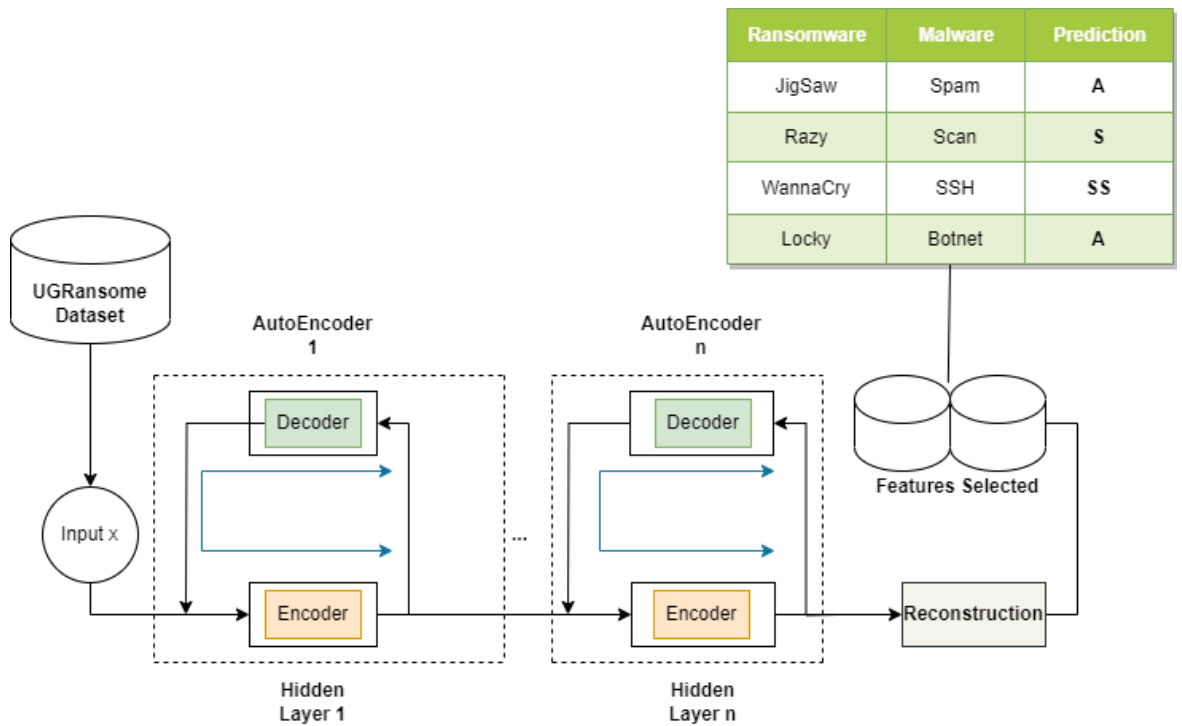

Fig.1: SAE Architecture

### 5.4 Unsupervised Pre-training Objective Function for Layer l

The objective function for unsupervised pre-training of layer l is given in Equation 1:

$$\min_{W^{(l)}, b^{(l)}} \frac{1}{m} \sum_{i=1}^{m} \left( x^{(l)}(i) - \hat{x}^{(l)}(i) \right)^2 \tag{1}$$

Where:

$W^{(l)}$ - weights of layer l
$b^{(l)}$ - biases of layer l
$x^{(l)}(i)$ - input data for the i-th training example in layer l
$\hat{x}^{(l)}(i)$ - reconstructed output for the i-th training example in layer l
$m$ - number of training examples

### 5.5 Supervised Fine-Tuning

After unsupervised pre-training, the layers are stacked together to form the full SAE. The network is then trained using a supervised loss function, typically a classification loss, with labeled data [35]. The mathematical formulation of the supervised cross-entropy loss function is given in Equation 2:

$$L_{supervised} = -\frac{1}{N} \sum_{i=1}^{N} \sum_{j=1}^{C} y_{ij} \log(p_{ij}) \tag{2}$$

Where:

$L_{supervised}$ - Supervised loss
$N$ - Number of training examples
$C$ - Number of classes
$y_{ij}$ - Binary indicator (0 or 1) for correct classification
$p_{ij}$ - Predicted probability for example i to belong to class j

### 5.6 Recurrent Neural Network

A recurrent Neural Network (RNN) is a variation of the feedforward neural network (NN) that introduces a recurrent structure within the network [34,36] (Figure 2). While the feedforward NN comprises multiple layers with unidirectional connections, RNN establishes connections from each neuron to itself. This self connection mechanism allows RNN to retain previous inputs to potentially influence the network's output [36].

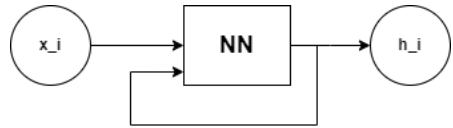

Fig.2: RNN Architecture

### 5.7 RNN Inference and Training

In RNN, the inference process is similar to that of the feedforward NN, completed through forward propagation. Training in RNN is accomplished using backpropagation through time, where the weights are updated based on the gradient [35]. However, RNN faces challenges such as the vanishing gradient problem and the exploding gradient problem. The gradient for each output in RNN depends not only on the current layer but also on the previous layer. Continuous updates of backpropagation can lead to vanishing gradients, causing weakening gradients. Conversely, when gradients become too large, it results in the exploding gradient problem [36].

### 5.8 RNN and LSTM

To address RNN issues, the LSTM deep learning algorithm was developed by Hochreiter and Schmidhuber in 1997 as a variant of the RNN model [35, 36]. LSTM introduces the concept of memory cells for its nodes to enable the linkage of prior data information to the present nodes. Each LSTM node incorporates three gating mechanisms: an input gate, a forget gate, and an output gate (Figure 3).

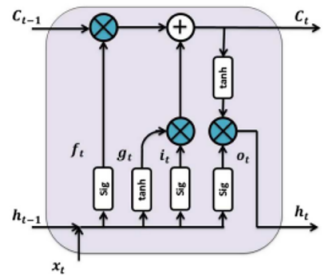

Fig.3: LSTM Node with Gating Mechanisms [34,36]

The key components of the LSTM gating mechanisms can be defined as follows:

– $i_t$ **(Input Gate)**: Controls the flow of new information into the memory cell.

- **$f_t$ (Forget Gate)**: Controls the flow of information to forget from the previous memory cell state.
- **$o_t$ (Output Gate)**: Controls the output from the memory cell.
- **$c_t$**: Represents the cell state.
- **$h_t$**: Represents the hidden state.

The LSTM equations for these gating mechanisms are as follows:

**Input Gate**:
$$i_t = \sigma(W_i \times [h_{t-1}, x_t] + b_i) \tag{3}$$

**Forget Gate**:
$$f_t = \sigma(W_f \times [h_{t-1}, x_t] + b_f) \tag{4}$$

**Output Gate**:
$$o_t = \sigma(W_o \times [h_{t-1}, x_t] + b_o) \tag{5}$$

**Cell State Update**:
$$c_t = f_t \times c_{t-1} + i_t \times \tanh(W_c \times [h_{t-1}, x_t] + b_c) \tag{6}$$

**Hidden State Update**:
$$h_t = o_t \times \tanh(c_t) \tag{7}$$

Where: - $W_i, W_f, W_o, W_c$ are weight matrices for the gates. - $b_i, b_f, b_o, b_c$ are bias vectors for the gates. - $\sigma$ represents the sigmoid activation function. -tanh represents the hyperbolic tangent activation function. - $[h_{t-1}, x_t]$ represents the concatenation of the previous hidden state $h_{t-1}$ and the current input $x_t$. These equations govern the behavior of the LSTM memory cell and its gating mechanisms, allowing it to capture long-term dependencies in sequential data [35].

### 5.9 Extreme Gradient Boosting

In this research, we compare the performance of the proposed SAE-LSTM model with that of the XGBoost algorithm. We used the UGRansome dataset for evaluation. XGBoost (Extreme Gradient Boosting) is a powerful and efficient machine learning algorithm used for both regression and classification tasks [37].

It belongs to the ensemble learning category and is based on the gradient boosting framework. XGBoost is known for its high predictive accuracy and is widely used in various data science and machine learning competitions. XGBoost aims to find an optimal model by minimizing a loss function that measures the difference between predicted values and actual target values [37]. The algorithm builds a strong predictive model by combining multiple weak models (decision trees) iteratively.

This algorithm uses the following concepts:

**Objective function**:
$$obj(\theta) = L(\theta) + \Omega(\theta) \tag{8}$$

This is the overall function that XGBoost aims to optimize during training [37]. It is a combination of two main parts: the **loss function (L(θ))** and the **regularization** term (Ω(θ)). The goal of XGBoost is to find the best values of model parameters that minimize this objective function [37].

**Loss function L(θ)**:

$$L(\theta) = \sum_{i=1}^{n} l(y_i, \hat{y}_i^{(t)}) \tag{9}$$

This term measures the discrepancy between the **actual target values (y_i)** and the **predicted values (y^(t))** generated by the current iteration of the model [37]. The loss function quantifies how well the model is performing on the training data. The objective is to minimize this loss by adjusting the model's parameters [37].

**Regularization term Ω(θ)**:

$$\Omega(\theta) = \gamma T + \frac{1}{2}\lambda \sum_{j=1}^{T} w_j^2 \tag{10}$$

Regularization is a technique used to prevent overfitting, which occurs when a model fits the training data too closely and does not generalize well to new data [37]. In XGBoost, there are two components to the regularization term:

γT : It discourages the model from creating too many complex rules.  (11)

$\frac{1}{2}\lambda \sum w_j^2$ : Discourages the model from assigning excessively large weights.

(12)

**Prediction function**:

$$\hat{y}_i^{(t)} = \phi(x_i) = \sum_{k=1}^{K} f_k(x_i) \tag{13}$$

This function computes the predicted value for a specific data point (x_i) at a given iteration (t) of the boosting process [37]. It is essentially the sum of predictions from individual trees (f_k(x_i)) in the model. As boosting iterations progress, more trees are added, and the prediction is updated. In summary, XGBoost seeks to find the best model parameters (θ) by minimizing a combination of two factors: how well the model fits the training data (loss function) and how complex the model is (regularization term).

The prediction function (y^(t)) represents the model's output for a specific data point at a given iteration of boosting. The goal is to iteratively improve the model by adjusting its parameters and thereby reducing the overall objective function.

## 5.10    Performance Evaluation

The evaluation of the training and testing performance of the established models for ransomware classification is crucial. Several metrics are commonly used to assess the effectiveness of these models, including accuracy, precision, recall (sensitivity), and the F1 Score [12,38,39]. These metrics provide valuable insights into the model's ability to make accurate predictions.

## 5.15    Confusion Matrix

A confusion matrix is often used to provide a more detailed evaluation of model performance [12]. The methodological approach employed in this research is visually depicted in Algorithm 1.

---

**Algorithm 1 SAE-LSTM Training**

---

1: Initialize the encoder and decoder neural network models

2: Define a loss function L (e.g., Mean Squared Error)

3: Define an optimizer (e.g., Stochastic Gradient Descent)

4: for each training epoch do

5:        for each training batch do

6:              Forward pass:

7:              Pass the input data $x_t$ through the encoder to obtain encoded features $h_t = \sigma(W_i \times [h_{t-1}, x_t] + b_i)$

8:              Pass the encoded features $h_t$ through the decoder to obtain decoded features $\hat{x}_t = \sigma(W_o \times h_t + b_o)$

9:              Compute the loss: $L(x_t, \hat{x}_t)$

10:             Backward pass:

11:             Calculate gradients using backpropagation

12:             Update the weights and biases $W_i$, $W_o$, $b_i$, $b_o$ using the optimizer

13:         Compute the average loss for the epoch

14:         if the average loss is below a predefined threshold or after a fixed number of epochs then

15:              break                                        ▷ Training convergence criteria met

16: Use the trained autoencoder for feature selection:

17: Extract the encoded features $h_t$ from the encoder

18: These encoded features can be used as selected features for LSTM classification

## 6    Experimental Setups

In this investigation, both the training and testing phases of the proposed data preprocessing, feature extraction, and classification models were executed using Python programming language, version 3.10.12. The training and testing phases of the proposed data encoding, normalization, SAE, and LSTM model were carried out using the Google Colaboratory cloud system. This platform offers convenient access to a wide array of Python libraries and services at no cost. To enhance algorithm execution speed, Nvidia CUDA technology within the Colab environment was utilized.

Various essential tasks, including file uploading, data preprocessing, data frame setup, and more, were accomplished using Python libraries such as numpy, pandas, statistics, sklearn, matplotlib.pyplot, and seaborn.

For implementing the recommended SAE and LSTM architecture, the Python TensorFlow Keras library was employed. The specified SAE architecture comprised three encoders with 75, 50, and 13 layers, respectively, and three corresponding decoders with 50, 75, and 13 layers (Table 3). The activation function was configured as relu, the optimizer as Adam, the loss as mean squared error (mse), and the number of epochs as 50 (Tables 3). The constructed LSTM network consisted of 3 layers, each containing 168 neurons (Table 4). The loss parameter was set to sparse categorical cross-entropy, the optimizer to Adam, and the number of epochs to 400.

Table 3: SAE Layers and Parameters

| Layer (type) | Output Shape | Param # |
|---|---|---|
| input_1 (InputLayer) | (None, 13) | 0 |
| dense | (None, 75) | 1,050 |
| dense_1 | (None, 50) | 3,800 |
| dense_2 | (None, 13) | 663 |
| dense_3 | (None, 50) | 700 |
| dense_4 | (None, 75) | 3,825 |
| dense_5 | (None, 13) | 988 |
| Total params: | | 11,026 |

Table 4: LSTM Layers and Parameters

| Layer (type) | Output Shape | Param # |
|---|---|---|
| lstm_3 (LSTM) | (None, 168) | 122,304 |
| dense_21 (Dense) | (None, 3) | 507 |
| Total params: | | 122,811 |

## 7 Results

In this section, we delve into the outcomes obtained through our proposed computational framework. We provide a comprehensive discussion of various facets, including the data preprocessing and encoding procedures, the results of feature extraction utilizing SAE, the cross-validation process involving data splitting, the performance of the LSTM classification, and the predictive modeling of ransomware, categorizing it into **Signature (S), Anomaly (A), and Synthetic Signature (SS)**.

### 7.1 Data Encoding and Pre-processing

Figure 6 provides an overview of the UGRansome statistics. The original UGRansome consists of 207,533 features, with 58,491 redundant patterns that account for 28.18% of the dataset (Figure 6). Within the scope of this study, the sklearn preprocessing library played a pivotal role in the conversion of categorical attributes into numeric representations across multiple columns within the UGRansome dataset (Figure 7). To eliminate redundancy, the SAE ignored duplicate rows during the feature selection process (Figure 6). We employed a methodology known as label encoding to transform UGRansome data. The primary objective underlying this encoding strategy was to render the dataset compatible with machine learning algorithms that mandate numeric inputs for their operation. By this process, categorical variables were effectively transformed into numerical equivalents, thereby rendering them amenable to various modeling and analytical techniques.

**Dataset Statistics**

| | |
|---|---|
| Number of Variables | 14 |
| Number of Rows | 207533 |
| Missing Cells | 0 |
| Missing Cells (%) | 0.0% |
| Duplicate Rows | 58491 |
| Duplicate Rows (%) | 28.2% |
| Total Size in Memory | 106.9 MB |
| Average Row Size in Memory | 540.2 B |
| Variable Types | Numerical: 4<br>Categorical: 9<br>GeoGraphy: 1 |

(a) Dataset with redundancy

**Dataset Statistics**

| | |
|---|---|
| Number of Variables | 14 |
| Number of Rows | 149042 |
| Missing Cells | 0 |
| Missing Cells (%) | 0.0% |
| Duplicate Rows | 0 |
| Duplicate Rows (%) | 0.0% |
| Total Size in Memory | 78.0 MB |
| Average Row Size in Memory | 548.5 B |
| Variable Types | Numerical: 4<br>Categorical: 9<br>GeoGraphy: 1 |

(b) Dataset without redundancy

Fig.6: Dataset Characteristics

|  | Time | Protocol | Flag | Ransomware | Clusters | SeedAddress | ExpAddress | BTC | USD | Netflow_Bytes | IPaddress | Malware | Port | Prediction |
|---|---|---|---|---|---|---|---|---|---|---|---|---|---|---|
| 0 | 40 | TCP | A | WannaCry | 1 | 1DA11mPS | 1BonuSr7 | 1 | 504 | 8 | A | Bonet | 5061 | SS |
| 1 | 30 | TCP | A | WannaCry | 1 | 1DA11mPS | 1BonuSr7 | 1 | 508 | 7 | A | Bonet | 5061 | SS |
| 2 | 20 | TCP | A | WannaCry | 1 | 1DA11mPS | 1BonuSr7 | 1 | 512 | 15 | A | Bonet | 5061 | SS |
| 3 | 57 | TCP | A | WannaCry | 1 | 1DA11mPS | 1BonuSr7 | 1 | 516 | 9 | A | Bonet | 5061 | SS |
| 4 | 41 | TCP | A | WannaCry | 1 | 1DA11mPS | 1BonuSr7 | 1 | 520 | 17 | A | Bonet | 5061 | SS |
| ... | ... | ... | ... | ... | ... | ... | ... | ... | ... | ... | ... | ... | ... | ... |
| 149037 | 33 | UDP | AP | TowerWeb | 3 | 1AEoiHYZ | 1SYSTEMQ | 1010 | 1590 | 3340 | A | Scan | 5062 | A |
| 149038 | 33 | UDP | AP | TowerWeb | 3 | 1AEoiHYZ | 1SYSTEMQ | 1014 | 1596 | 3351 | A | Scan | 5062 | A |
| 149039 | 33 | UDP | AP | TowerWeb | 3 | 1AEoiHYZ | 1SYSTEMQ | 1018 | 1602 | 3362 | A | Scan | 5062 | A |
| 149040 | 33 | UDP | AP | TowerWeb | 3 | 1AEoiHYZ | 1SYSTEMQ | 1022 | 1608 | 3373 | A | Scan | 5062 | A |
| 149041 | 33 | UDP | AP | TowerWeb | 3 | 1AEoiHYZ | 1SYSTEMQ | 1026 | 1614 | 3384 | A | Scan | 5062 | A |

149042 rows × 14 columns

(a) The original dataset

|  | Time | Protocol | Flag | Ransomware | Clusters | SeedAddress | ExpAddress | BTC | USD | Netflow_Bytes | IPaddress | Malware | Port | Prediction |
|---|---|---|---|---|---|---|---|---|---|---|---|---|---|---|
| 0 | 40 | 1 | 0 | 16 | 1 | 2 | 2 | 1 | 504 | 8 | 0 | 1 | 5061 | 2 |
| 1 | 30 | 1 | 0 | 16 | 1 | 2 | 2 | 1 | 508 | 7 | 0 | 1 | 5061 | 2 |
| 2 | 20 | 1 | 0 | 16 | 1 | 2 | 2 | 1 | 512 | 15 | 0 | 1 | 5061 | 2 |
| 3 | 57 | 1 | 0 | 16 | 1 | 2 | 2 | 1 | 516 | 9 | 0 | 1 | 5061 | 2 |
| 4 | 41 | 1 | 0 | 16 | 1 | 2 | 2 | 1 | 520 | 17 | 0 | 1 | 5061 | 2 |
| ... | ... | ... | ... | ... | ... | ... | ... | ... | ... | ... | ... | ... | ... | ... |
| 149037 | 33 | 2 | 2 | 15 | 3 | 1 | 1 | 6 | 1010 | 1590 | 3340 | 0 | 6 | 5062 | 0 |
| 149038 | 33 | 2 | 2 | 15 | 3 | 1 | 1 | 6 | 1014 | 1596 | 3351 | 0 | 6 | 5062 | 0 |
| 149039 | 33 | 2 | 2 | 15 | 3 | 1 | 1 | 6 | 1018 | 1602 | 3362 | 0 | 6 | 5062 | 0 |
| 149040 | 33 | 2 | 2 | 15 | 3 | 1 | 1 | 6 | 1022 | 1608 | 3373 | 0 | 6 | 5062 | 0 |
| 149041 | 33 | 2 | 2 | 15 | 3 | 1 | 1 | 6 | 1026 | 1614 | 3384 | 0 | 6 | 5062 | 0 |

149042 rows × 14 columns

(b) The preprocessed/encoded dataset

Fig.7: Comparison of Characteristics in the Pre-processed Dataset

## 7.2    Feature Selection Based SAE Results for Ransomware Classification

The initial phase of the analysis involved an examination of the distribution of ransomware instances selected by the SAE. It was observed that Locky, SamSam, and WannaCry exhibited the highest frequency of occurrences, whereas EDA2 and DMALocker occupied a middle ground, with NoobCrypt registering a relatively lower count (Figure 9).

Concurrently, an assessment of the cumulative costs associated with these ransomware types revealed that Locky, SamSam, and WannaCry still retained substantial monetary impact (Figure 10 (a)). Furthermore, an exploration of the distribution of various malware categories across ransomware types was conducted.

The results indicated a relatively balanced distribution, with SSH accounting for 33.0% of instances, Spam representing 31%, and UDP scan comprising 27.6%. In contrast, NerisBonet was found to be in the minority, constituting only 8.3% of the dataset (Figure 11). Subsequently, to gain a more standardized perspective and discern the true extent of the threat posed by each ransomware variant, an analysis of the average dollars per ransomware was undertaken. Surprisingly, this analysis yielded results divergent from the initial observation shown in Figure 10 (a). Locky, SamSam, and WannaCry did not occupy the top three positions in this ranking. Instead, NoobCrypt, previously positioned on the lower end of the frequency spectrum, emerged as a leading contender, joined by EDA2 and DMALocker, both previously situated within the middle range (Figure 10 (b)).

(a) Anomaly of timestamp

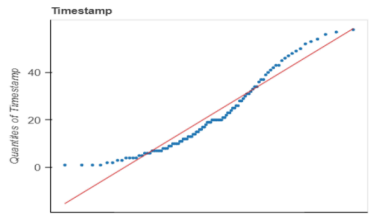

(b) Removal of abnormal timestamp

(b) Timestamp standard deviation

Fig.8: Removal of abnormal timestamp in the Pre-processed Dataset

This result provides valuable insights into the UGRansome dataset, illustrating that while Locky, SamSam, and WannaCry may have incurred substantial cumulative damages due to their higher volume of attacks (Figure 10 (a)), they may not inflict as much financial harm per individual attack when compared to NoobCrypt, DMALocker, and EDA2 (Figure 10 (b)). Therefore, the latter ransomware variants should be closely monitored as potential major threats, particularly if the volume of their attacks were to increase. A correlation matrix of the SAE assesses relationships between features (Figure 12), with +1 indicating a strong positive linear correlation, -1 suggesting a strong negative

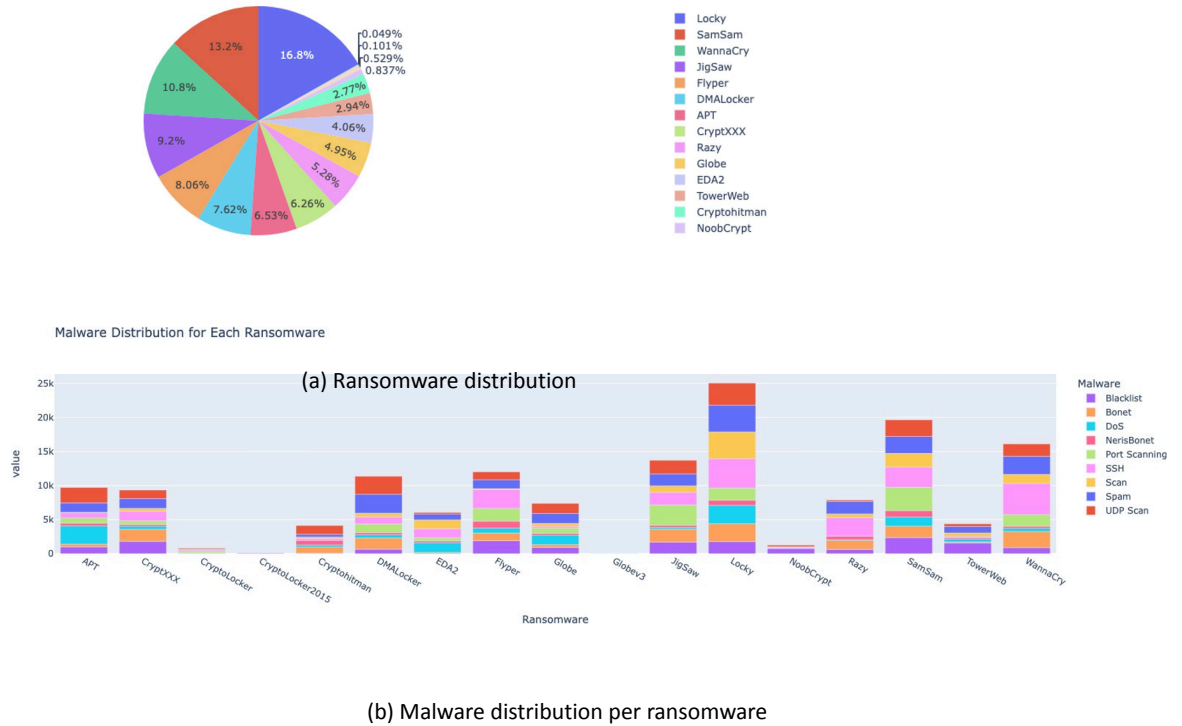

(a) Ransomware distribution

(b) Malware distribution per ransomware

Fig.9: Distribution of Attacks in the Pre-processed Dataset

correlation, and values near 0 denoting weak or no correlation. It helps identify correlated features, interpret their impact on models, and guide feature selection. Visualizing it via a heatmap enhances pattern recognition by color coding high positive, high negative, and low correlations. The ransomware attacks in the dataset inflicted severe financial devastation. On average, victims paid a staggering 30.69 BTC, equivalent to $798,602 (USD) as of September 2023, with an average dollar payout of $14,873.43 (USD). Figure 10 underscores the substantial financial toll imposed by ransomware threats. In fact, this aligns with previous findings where 11% of organizations that opted to pay ransoms in a 2021 report disclosed payments of $1 million or higher.[1] The average network traffic observed was 2021.16 bytes, with a considerable standard deviation of 2272.54 (Figure 13 (b)). This suggests a notable variation in values, potentially indicating spikes in network traffic triggered by zero-day threats. To address this, additional feature engineering might be necessary to better balance the dataset. Figure 13 (a) shows that CryptoLocker exhibited the most anomalous behaviors among ransomware. This suggests that zero-day threats like CryptoLocker, which restrict users' access to their computers, exhibit highly deviant behaviors

---

[1] https://news.sophos.com/en-us/2022/04/27/the-state-of-ransomware-2022/

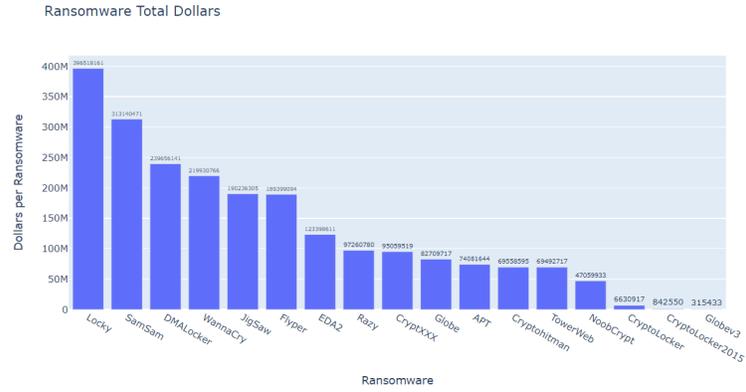

(a) Ransomware damage in USD

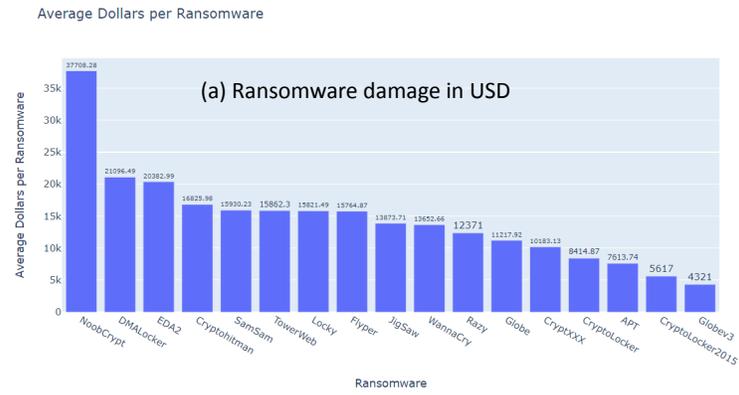

(b) Average ransomware damage in USD

Fig.10: Financial Damages of Ransomware in the Pre-processed Dataset

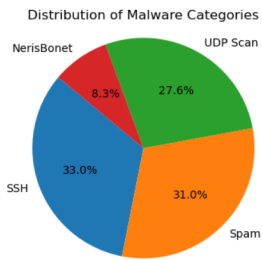

Fig.11: Distribution of Malware in the Pre-processed Dataset

compared to normal patterns, possibly due to their technical complexity and social engineering tactics [41].

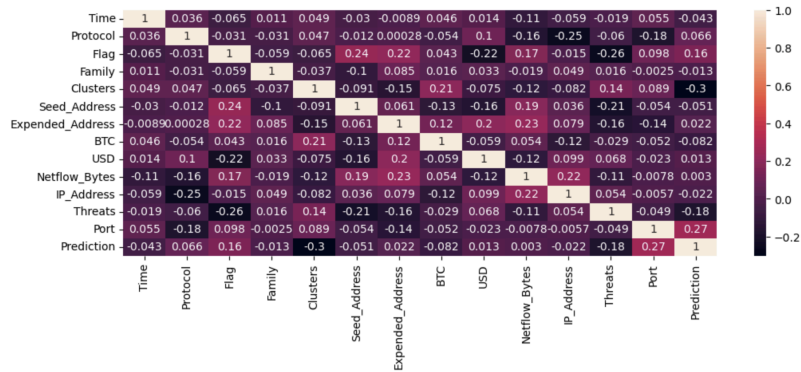

Fig.12: Correlation Matrix of SAE

### 7.3   SAE-Based LSTM Model Results

The LSTM model's classification outcomes, as shown in Figure 15 and Table 5, are detailed using a confusion matrix. The matrix highlights that 17,891 instances of ransomware were correctly classified as Signature (S) types, with over 11,000 instances classified as Synthetic Signature (SS) and Anomaly (A) types. This classification has an average accuracy of 98% (Figure 14 and Table 5). In our investigation, we undertook a comprehensive comparison between our novel SAE-LSTM model and an XGBoost algorithm described in [7]. The results obtained from the XGBoost algorithm, employing the UGRansome dataset [7], have been thoughtfully summarized in Table 6. Interestingly, this analysis revealed the superior performance of the SAE-LSTM model over the XGBoost algorithm, which can be attributed to the effectiveness of feature selection inherent to the SAE-LSTM approach (Figure 16).

Table 5: SAE-LSTM Performance Metrics

|          | Precision | Recall   | F1 Score | Total Features |
|----------|-----------|----------|----------|----------------|
| A        | 0.971879  | 0.986131 | 0.978953 | 11,320 S       |
|          | 0.991466  | 0.978024 | 0.984699 | 18,293 SS      |
|          | 0.987558  | 0.994367 | 0.990951 | 11,894         |
| Accuracy |           |          | 0.984918 | 41,507         |
| Average  | 0.985004  | 0.984918 | 0.984924 | 41,507         |

## 8   Discussion

In summary, Table 5 and Figure 14 demonstrate the robust performance of the SAE-based LSTM model across three distinct attack categories: Anomaly (A),

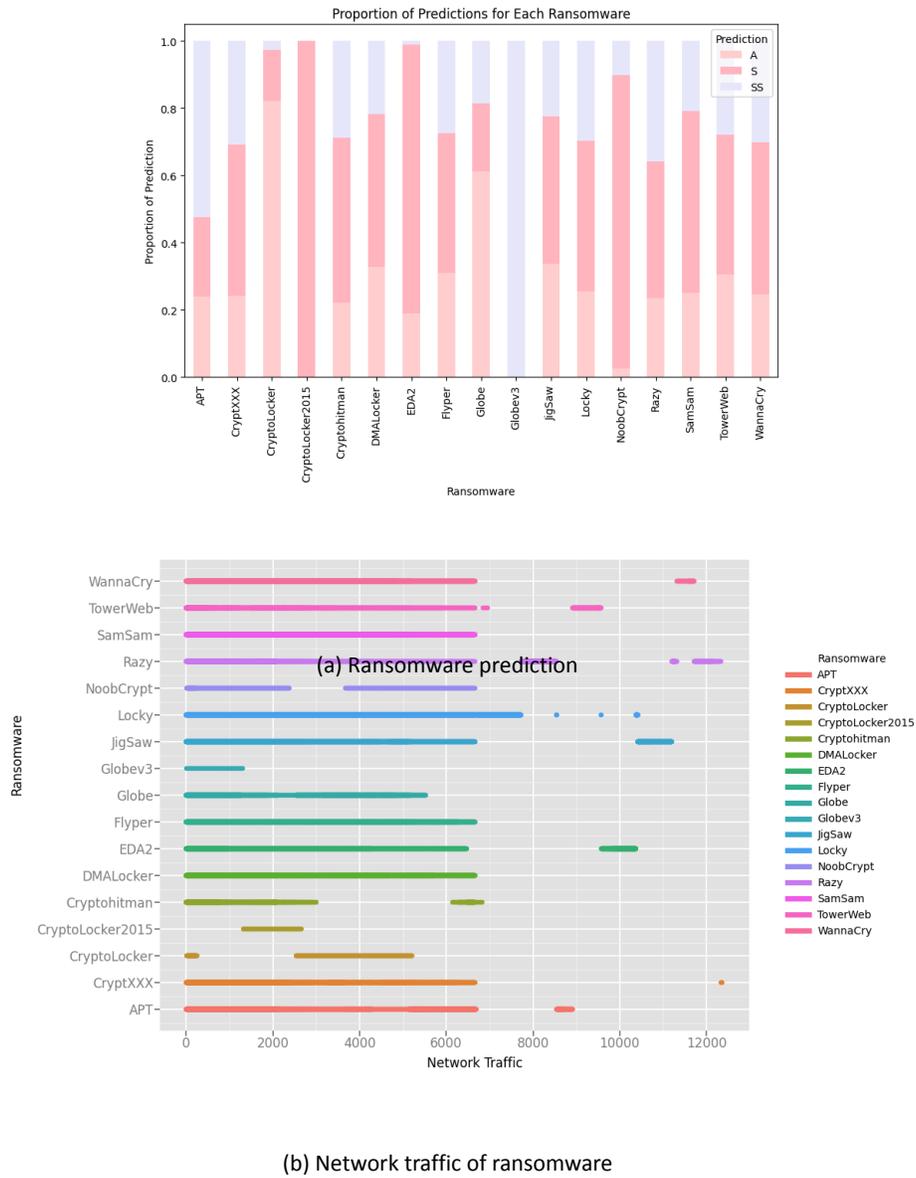

(a) Ransomware prediction

(b) Network traffic of ransomware

Fig.13: Ransomware Prediction and Network Traffic

Signature (S), and Synthetic Signature (SS). The model exhibits high precision, recall, and F1-score values, underscoring its effectiveness in accurately identifying various attack types. Moreover, the balanced average scores imply that the model generalizes well across different attack categories, ensuring consistent performance. The confusion matrix, depicted in figures 15 and 16, provides a visual representation of the model's performance, showcasing the number of true positive (TP), true negative (TN), false positive (FP), and false negative (FN) predictions. The proposed model demonstrates outstanding performance in identifying signature attacks, achieving a 98% accuracy rate, showcasing its proficiency in recognizing well established ransomware patterns such as **Locky, JigSaw, and WannaCry**. This outperforms the XGBoost model, which achieved a 95% accuracy rate in the same task (Figure 17). However, its slightly lower performance in identifying synthetic signature attacks highlights the challenge of detecting zero-day attack signatures. Anomaly attacks, representing novel threats, present a greater challenge due to their lack of discernible patterns. Future work in the IDS field could leverage the UGRansome dataset and refine model parameters to enhance anomaly detection. Table 7 provides a comparative analysis of various studies in the field of IDS.

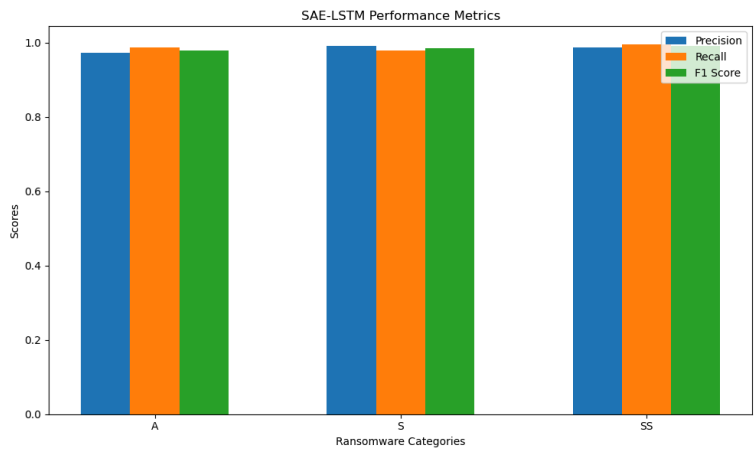

Fig.14: Overall performance of the SAE-LSTM Model

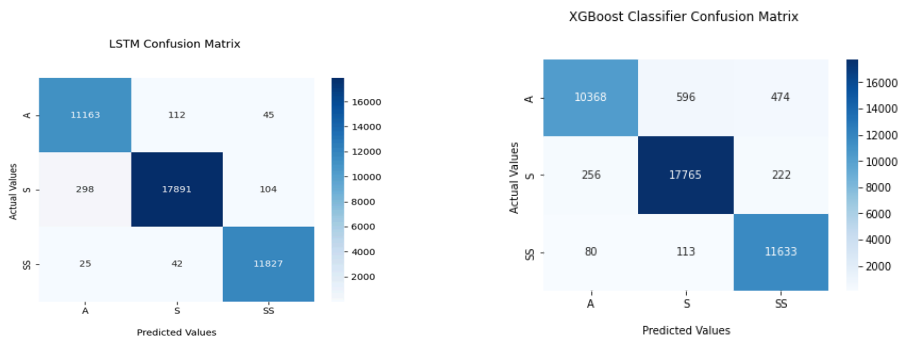

(a) SAE-LSTM                          (b) XGBoost

Fig.15: SAE-LSTM vs. XGBoost: Performance Analysis

Table 6: XGBoost Classifier Results

|   | Precision | Recall | F1-Score | Total Features |
|---|-----------|--------|----------|----------------|
| A | 0.9036    | 0.9314 | 0.9609   | 11,436         |
| S | 0.9585    | 0.9728 | 0.9656   | 18,249         |
| SS| 0.9447    | 0.9776 | 0.9609   | 11,822         |
| Accuracy | | | 0.9551 | |
| Macro Avg | 0.9547 | 0.9513 | 0.9526 | 41,507 |
| Weighted Avg | 0.9553 | 0.9551 | 0.9548 | 41,507 |

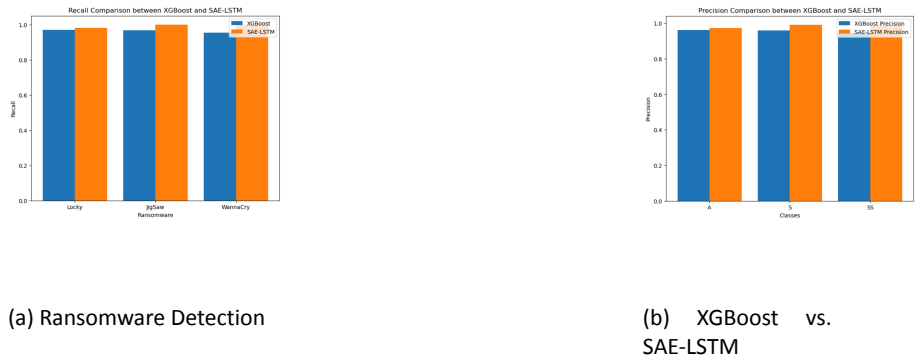

(a) Ransomware Detection                    (b)  XGBoost  vs. SAE-LSTM

Fig.16: Comparison Between XGBoost and SAE-LSTM

While many studies have achieved high accuracy, there are several limitations. Some limitations include the use of **shallow learning architectures**, scalability issues, domain-specific focus, and the need for labeled datasets. Our proposed research achieved a remarkable accuracy of 99% in ransomware classification using SAE and LSTM (Figure 17), but it is limited to supervised ransomware classification, and further research is required to assess its applicability to a broader range of intrusion detection scenarios. Figure 18 shows that the TP and TN of the proposed SAE-LSTM are higher, while FP and FN of the XGBoost are lower, hence the SAE-LSTM model is better in terms of overall classification accuracy and error rates compared to the XGBoost model (Figure 19). Figure 18 indicates that the SAE-LSTM model is more reliable in making correct predictions and has a higher precision and recall. Therefore, in this scenario, the SAE-LSTM model is considered better for ransomware detection.

## 9   Conclusion

In today's digital landscape, ransomware presents a formidable threat to individuals and businesses, prompting our innovative approach to detection and classification. Our method combines an SAE for feature selection with an LSTM classifier, yielding enhanced precision in categorizing ransomware. This process involves the UGRansome dataset preprocessing, unsupervised SAE feature selection, and supervised fine-tuning, resulting in a robust model that excels across diverse ransomware families. Architectural optimizations culminate in an exceptional 99% accuracy, surpassing conventional classifiers.

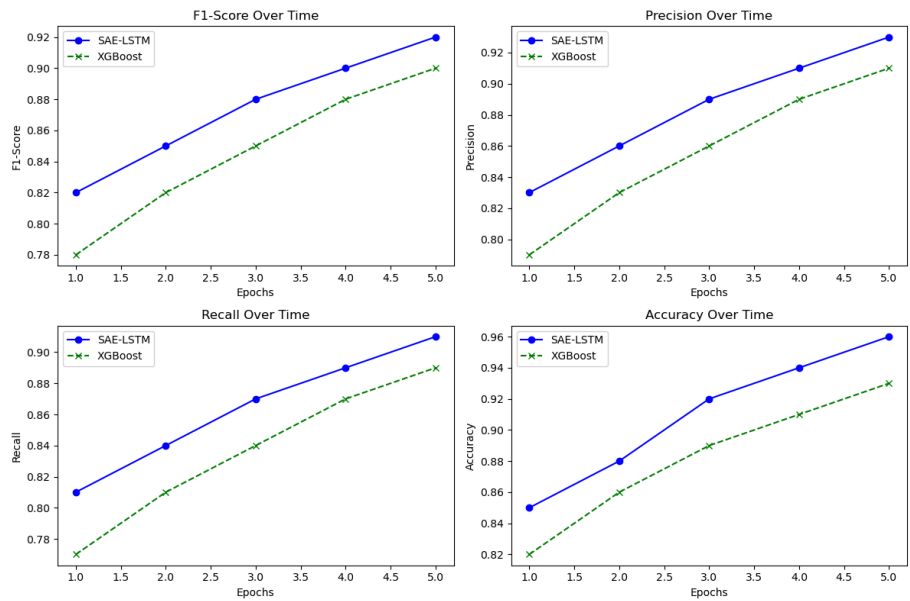

(a) Model performance

Fig.17: Performance Analysis

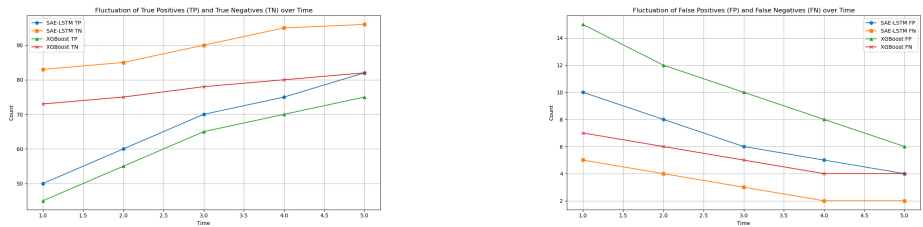

(a) TP vs. TN  (b) FP vs. FN

Fig.18: Classification Comparison Between XGBoost and SAE-LSTM